\documentclass[twocolumn,letterpaper]{IEEEAerospaceCLS}  

\usepackage[hidelinks]{hyperref}       
\usepackage{url}            
\usepackage{amsfonts}       
\usepackage{nicefrac}       
\usepackage{microtype}      
\usepackage{siunitx}
\usepackage{subfig}
\usepackage[]{xcolor}

\usepackage[]{graphicx}    
\newcommand{\ignore}[1]{}  

\newcommand{\ddist}{
    \begin{table}[h]
    \centering
    \caption{\textbf{Semantic space and distribution}}
    \scalebox{0.8}{
        \begin{tabular}{rlccc}
            & & \textbf{\% pixels} & \textbf{\% cases} & \textbf{Mask color} \\
            \hline
            0.&None / background       & 93.95\% & 100\% & \colorbox[rgb]{0,0,0}{\strut~~~~} \\
            1.&Solar panels            & 1.75\% & 100\% & \colorbox[rgb]{1,0,0}{\strut~~~~}\\
            2.&Solar panel drive shaft & 0.03\% & 49.99\% & \colorbox[rgb]{0,1,0}{\strut~~~~}\\
            3.&Antenna                 & 0.05\% & 73.30\% & \colorbox[rgb]{0,0,1}{\strut~~~~}\\
            4.&Parabolic reflector     & 0.08\% & 25.00\% & \colorbox[rgb]{1,1,0}{\strut~~~~}\\
            5.&Main module             & 2.90\% & 100\% & \colorbox[rgb]{1,0,1}{\strut~~~~}\\
            6.&Telescope               & 0.58\% & 25.00\% & \colorbox[rgb]{0,1,1}{\strut~~~~}\\
            7.&Main thrusters          & 0.05\% & 77.03\% & \colorbox[rgb]{1,0.5,0.5}{\strut~~~~}\\
            8.&Rotational thrusters    & 0.02\% & 99.99\% & \colorbox[rgb]{1,0.8,0.3}{\strut~~~~}\\
            9.&Sensors                 & 0.53\% & 98.27\% & \colorbox[rgb]{0.5,1,1}{\strut~~~~}\\
           10.&Launch vehicle adapter  & 0.08\% & 49.44\% & \colorbox[rgb]{0.25,0.5,0.75}{\strut~~~~}\\
        \hline
        \end{tabular}
    }
    \label{tab:ddist}
    \end{table}
}

\newcommand{\bigresults}{
    \begin{table*}[t]
    \centering
    \scalebox{0.9}{
        \begin{tabular}{lllccc}
             & & & \multicolumn{3}{c}{\textbf{Sørensen–Dice coefficient}} \\
            \cline{4-6}
            \textbf{Model architecture} & \textbf{Backbone} & \textbf{Loss function} & \textbf{Validation} & \textbf{Test; known target} & 
                \textbf{Test; unknown target} \\
            \hline
            DeepLab & ResNet50   & CCE          & 0.7703 & 0.7689          & \textbf{0.2519} \\
                    &            & Dice         & 0.7597 & 0.7593          & 0.2449          \\
                    &            & Dice + focal & 0.7869 & 0.7871          & 0.2502          \\
            HRNet   & HRNet\_w30 & CCE          & 0.7618 & 0.7618          & 0.2342          \\
                    &            & Dice         & 0.7712 & 0.8043          & 0.2404          \\
                    &            & Dice + focal & 0.7878 & 0.7886          & 0.2371          \\ 
            U-Net    & ResNet34   & CCE          & 0.8707 & \textbf{0.8723} & 0.2282          \\
                    &            & Dice         & 0.4423 & 0.4422          & 0.2284          \\
                    &            & Dice + focal & 0.8389 & 0.8395          & 0.2357          \\
            \hline
        \end{tabular}
    }
    \caption{\textbf{Test statistic (Sørensen–Dice coefficient) evaluated across experiments.  The consistent gap in evaluation criterion between known and unknown target tests suggest that memorization of components and configurations of the spacecraft present in both the training and known target test sets is occurring.}}
    \label{tab:big_results}
    \end{table*}
}

\newcommand{\perclassresults}{
    \begin{table*}[t]
    \centering
    \scalebox{0.9}{
        \begin{tabular}{rlccc}
            & & \multicolumn{2}{c}{\textbf{Sørensen–Dice coefficient}} \\
            \cline{3-4}
            \multicolumn{2}{c}{\textbf{Class}} & \textbf{Test; in-distribution} & \textbf{Test; out-of-distribution} \\
            \hline
            0.&None / background       & 0.9991 & 0.9972 \\
            1.&Solar panels            & 0.9836 & 0.7673 \\
            2.&Solar panel drive shaft & 0.7142 & 0.0259 \\
            3.&Antenna                 & 0.6481 & 0.0003 \\
            4.&Parabolic reflector     & 0.9493 & 0.0009 \\
            5.&Main module             & 0.9838 & 0.6678 \\
            6.&Telescope               & 0.9875 & NA     \\
            7.&Main thrusters          & 0.8707 & 0.0100 \\
            8.&Rotational thrusters    & 0.6404 & 0.0046 \\
            9.&Sensors                 & 0.9486 & 0.0131 \\
            10.&Launch vehicle adapter  & 0.8699 & NA    \\
            \hline
        \end{tabular}
    }
    \caption{\textbf{Sørenson-Dice coefficient by class for the best performing segmentation model at a test with a known target; a U-Net model with categorical cross-entropy loss.  The unknown target spacecraft, Clementine, lacks a telescope (category 6) and a launch vehicle adapter similar to the spacecraft in the known target training set, instead having an inter-stage adapter that was not given its own class label.}}
    \label{tab:perclass}
    \end{table*}
}

\newcommand{\practicalfigs}{
    \begin{figure*}[h]
        \centering
        \subfloat[Test image]{\includegraphics[width=0.31\textwidth]{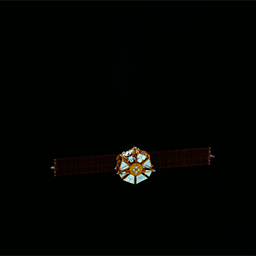}\label{fig:pract-im-0}}
        \hfill
        \subfloat[True mask]{\includegraphics[width=0.31\textwidth]{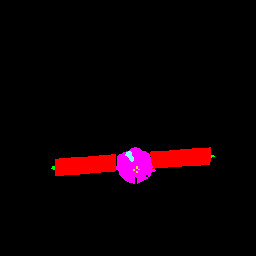}\label{fig:pract-true-mask-0}}
        \hfill
        \subfloat[Predicted mask]{\includegraphics[width=0.31\textwidth]{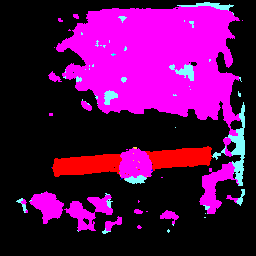}\label{fig:pract-pred-mask-0}}
        \hfill
        \subfloat[Test image]{\includegraphics[width=0.31\textwidth]{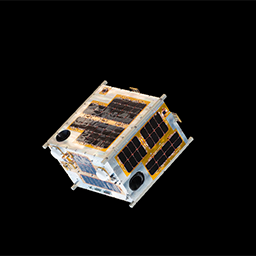}\label{fig:pract-im-1}}
        \hfill
        \subfloat[True mask]{\includegraphics[width=0.31\textwidth]{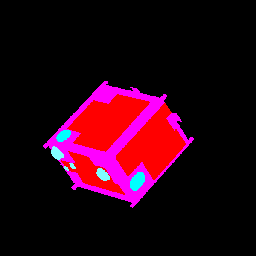}\label{fig:pract-true-mask-1}}
        \hfill
        \subfloat[Predicted mask]{\includegraphics[width=0.31\textwidth]{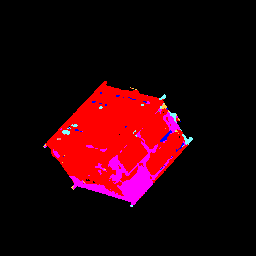}\label{fig:pract-pred-mask-1}}
        \hfill
        \subfloat[Test image]{\includegraphics[width=0.31\textwidth]{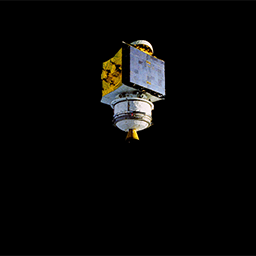}\label{fig:pract-im-2}}
        \hfill
        \subfloat[True mask]{\includegraphics[width=0.31\textwidth]{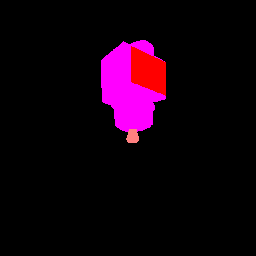}\label{fig:pract-true-mask-2}}
        \hfill
        \subfloat[Predicted mask]{\includegraphics[width=0.31\textwidth]{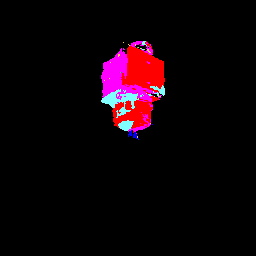}\label{fig:pract-pred-mask-2}}
        \caption{\textbf{True and predicted masks for three test images from the NASA Image and Video Library, with other spacecraft or planetary bodies removed from the original photo.  From top to bottom: the Space Flyer Unit (macro-average Sørensen–Dice coefficient: 0.2098), Diwata-1 (macro-average Sørensen–Dice coefficient: 0.2311), and Satcom K-2 (macro-average Sørensen–Dice coefficient: 0.2759)}}
        \label{fig:pract-test}
    \end{figure*}
}

\begin{document}
\title{Synthetic Data for Semantic Image Segmentation of Imagery of Unmanned Spacecraft}

\author{%
William S.~Armstrong\\ 
Stanford Center for Professional Development\\
Stanford University\\
Stanford, CA, 94305\\
wsa@nx-2.com
\and 
Spencer Drakontaidis\\
Department of Computer Science\\
Stanford University\\
Stanford, CA, 94305\\
spencer@drakontaidis.com
\and
Nicholas Lui\\
Department of Statistics\\
Stanford University\\
Stanford, CA, 94305\\
niclui@stanford.edu
\thanks{\footnotesize 978-1-6654-9032-0/23/$\$31.00$ \copyright2023 IEEE}              
}

\maketitle

\thispagestyle{plain}
\pagestyle{plain}

\maketitle

\thispagestyle{plain}
\pagestyle{plain}

\begin{abstract}

Images of spacecraft photographed from other spacecraft operating in outer space are difficult to come by, especially at a scale typically required for deep learning tasks. Semantic image segmentation, object detection and localization, and pose estimation are well researched areas with powerful results for many applications, and would be very useful in autonomous spacecraft operation and rendezvous. However, recent studies show that these strong results in broad and common domains may generalize poorly even to specific industrial applications on earth. To address this, we propose a method for generating synthetic image data that are labelled for semantic segmentation, generalizable to other tasks, and provide a prototype synthetic image dataset consisting of 2D monocular images of unmanned spacecraft, in order to enable further research in the area of autonomous spacecraft rendezvous. We also present a strong benchmark result (Sørensen-Dice coefficient 0.8723) on these synthetic data, suggesting that it is feasible to train well-performing image segmentation models for this task, especially if the target spacecraft and its configuration are known.
\end{abstract}

\tableofcontents

\section{Introduction}

Images of spacecraft photographed from other spacecraft operating in outer space are difficult to come by, especially at a scale typically required for deep learning tasks.  Semantic image segmentation, object detection and localization, and pose estimation are well researched areas with powerful results for many applications, and would be very useful in autonomous spacecraft operation and rendezvous.  However, Wong et al.~\cite{wong_synthetic_2019} note that these strong results in broad and common domains may generalize poorly even to specific industrial applications on earth.

To address this, we generated a prototype synthetic image dataset labelled for semantic segmentation of 2D images of unmanned spacecraft, and are endeavouring to train a performant deep learning image segmentation model using the same, with the ultimate goal of enabling further research in the area of autonomous spacecraft rendezvous.

\begin{figure}[h!]
    \centering
    \includegraphics[width=6cm]{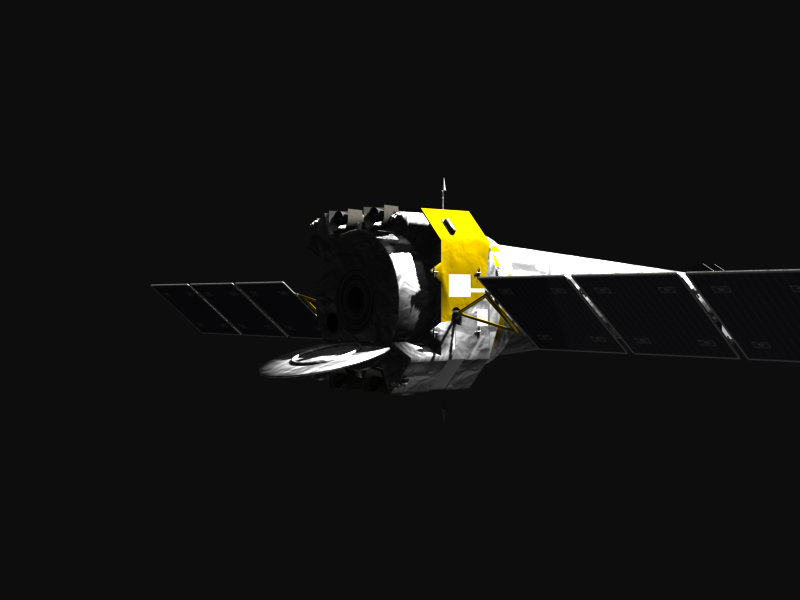}
    \caption{\textbf{Chandra spacecraft 3D Model}}
    \label{fig:chandra-image}
\end{figure}

\begin{figure}[h!]
    \centering
    \includegraphics[width=6cm]{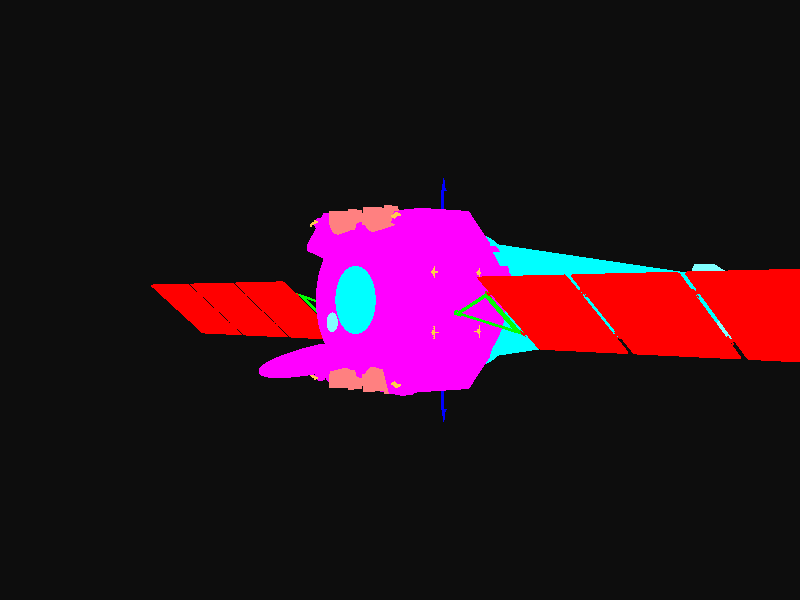}
    \caption{\textbf{Corresponding segmentation mask}}
    \label{fig:chandra-mask}
\end{figure}

\subsection{Related work}

Minaee et al.~\cite{minaee_image_2021} and Ghosh et al.~\cite{ghosh_understanding_2019} provide recent surveys of deep learning approaches for semantic image segmentation.  Treml et al.~\cite{treml_speeding_2016} discuss semantic segmentation using deep learning for autonomous terrestrial vehicles (i.e.~self-driving cars).  

Arantes et al.~\cite{arantes_far_2010} present a discussion of machine vision pose estimation for on-orbit autonomous satellite intercept and rendezvous with uncooperative spacecraft using a monocular optical sensor.  Kisantal et al.~\cite{kisantal_satellite_2020} present a synthetic dataset of images of unmanned spacecraft labelled for pose estimation, including images generated from 3D computer models, and hardware-in-the-loop simulation using a physical mock-up.

Wong et al.~\cite{wong_synthetic_2019} express concerns about the availability of training data at scale for domain-specific object recognition tasks, and propose a method of procedurally generating large image datasets from 3D models based on a small number of physical examples, labelled for image classification.

Yang et al.~\cite{yang_IEEEA_semantic} present an analysis of the impact of atmospheric distortion on performance of a semantic segmentation task on ground-based images of satellites on low earth orbit; the formulation of the segmentation task discussed herein largely follows that posed in this prior work, however, our aim is to present a method for generating pairs of synthetic images and ground truth labels suitable for semantic segmentation tasks in this domain, as well as provide some benchmark results using these data.  Our major hypothesis is that since this task is fundamentally about object or component recognition using data from an optical sensor, images with a degree of verisimilitude to the human eye would be effective for training deep learning models that would be useful in a practical application. 

To that end, we will describe the method by which the data were generated, discuss the baseline methods that were attempted and the evaluation criteria selected, and finally present the results of these along with a practical test of the best-performing method using three non-synthetic images of satellites taken from the vantage point of an on-orbit camera.

\section{Dataset}

The data generated consist of 60,000 renderings of one of four open-source 3D models of unmanned spacecraft published by NASA \cite{nasa_models_nodate}, with accompanying ground truth labels, created  using the open source 3D modeling software Blender \cite{blender}. 

\subsection{3D models}

To provide our dataset with a variety of spacecraft configurations, we chose the Chandra X-Ray Observatory, Near Earth Asteroid Rendezvous -- NEAR Shoemaker, Cluster II, and the IBEX Interstellar Boundary Explorer, as 3D models from which to generate training, validation, and test images in this domain.

We made some artistic modifications to the published 3D models in order to be able to generate more realistic looking images where possible, and attempted to simulate the lighting conditions of low-earth orbit by illuminating the 3D model with two light sources: one light source at infinity simulating the intensity, color, and parallel light rays of the sun, and one planar light source to simulate earthshine.  As such, our simulated environment assumes that rendezvous is taking place on the day side of the earth, which is perhaps not unreasonable since we are assuming an optical sensor.

We then duplicated the 3D model used for image generation and manually colored each polygon of the duplicate model from a discrete set of colors uniquely associated with one of the class labels.  Generation of the semantic masks was then accomplished by removing all light sources from the Blender scene and modifying the shaders, material properties, and renderer settings such that the ray-tracer projected only the appropriate color value onto each pixel of the rendered image based on the camera's perspective.

An example simulated image and ground truth mask can be seen in figures \ref{fig:chandra-image} and \ref{fig:chandra-mask}, respectively.

\subsection{Class labels}

We worked with an industry expert to define eleven class labels for the segmentation task, shown in table \ref{tab:ddist}, along with pixel- and case-level distributions of class prevalence among images generated from the four in-distribution spacecraft.

\ddist

Our objective was both to have a single meaningful semantic space that covered a variety of spacecrafts and configurations, and also for the class labels to provide a clear delineation between components to fixate on (e.g., the launch vehicle adapter) and components to avoid (e.g., thrusters) during rendezvous.

\subsection{Procedural image generation}

We used the Blender API for Python to automate our dataset generation in a scalable way.  A Python script moves the camera in a spherical pattern around the spacecraft to one of 5,000 positions.  For each position, three rendered images were generated with the same aspect, but with different ranges; one at the first position, and one each from approximately twice and three times the distance from the spacecraft, thus creating a total of 15,000 training images from each of the 3D models. The process was then repeated for the color-coded ground truth 3D model, taking corresponding renders from identical positions in the scene.  

\subsection{Ground truth representation}

The procedurally generated ground truth images were then post-processed using Python's Pillow library to represent a pixel-wise categorical encoding in the semantic space; we programatically assigned RGB color values in the generated images to their corresponding class labels, resulting in single-channel images with pixel values equal to the cardinal number associated with each category \cite{pillow}.  These single-channel PNG format images were generated with embedded color palettes for human and machine readability.

\subsection{Unknown target test set}\label{sec:ood}

In addition to the 60,000 images used for training, validation, and testing, we also generated an additional 1,500 images from a 3D model of a fifth spacecraft -- the Deep Space Program Science Experiment (DSPSE, a.k.a.~Clementine) --  to evaluate the ability of a semantic segmentation model trained on these data to identify specific components independently and generalize to an unknown target.

\section{Baseline methods}

We trained U-Net, HRNet, and DeepLab deep image segmentation models to determine which architecture performed the best for this task. All models were trained using Python's FastAI and SemTorch libraries \cite{fastai} \cite{semtorch} \cite{pytorch}, to enable rapid testing of a variety of benchmark models.  In each case, a backbone pre-trained on ImageNet \cite{imagenet} was incorporated to leverage transfer learning in extracting features from the input image. \cite{imagenet}.

\subsection{U-Net}

Ronneberger et al.~\cite{ronneberger_u-net_2015} describe U-Net, which aims to provide precise localization even when using a smaller dataset than is typically used for image segmentation tasks.  SemTorch \cite{fastai} uses PyTorch's \cite{pytorch} implementation of U-Net.

For the U-Net model we trained, we selected a ResNet34 backbone.  A batch size of 8 was chosen as the maximum feasible batch size for this model, given the GPU that was used.

\subsection{HRNet}

HRNet (High-Resolution Net) is a CNN developed specifically to retain and use high-resolution inputs throughout the network, resulting in better performing for pixel labelling and segmenation tasks \cite{hrnet}.  HRNet aims to provide high spatial precision, which is desirable in this task due to the variety of classes and class imbalance \cite{hrnet}.  

We selected a pre-trained HRNet30 backbone to perform feature extraction. A batch size of 16 was chosen using the same criteria as before.

\subsection{DeepLab}

DeepLab is a CNN developed and open-sourced by Google that relies on atrous convolutions to perform image segmentation tasks \cite{chen_encoder-decoder_2018}. More specifically, we used the latest iteration of the DeepLab model at time of writing, DeepLabv3+, as implemented by FastAI.  DeepLabv3+ aims to incorporate the best aspects of spatial pyramid pooling and encoder-decoder models that leads to a faster and more performant model overall \cite{chen_encoder-decoder_2018}.

In this case, a ResNet50 backbone was selected over ResNet34, due to limitations of the SemTorch library. A batch size of 16 was chosen by the same criteria as before.

\subsection{Loss functions}

We experimented with three different loss functions: categorical cross-entropy loss, Dice loss, and a mixture of focal and Dice losses.  These choices were motivated by the considerable class imbalance as shown in table \ref{tab:ddist}.

\subsubsection{Categorical cross-entropy loss}
For each pixel, this function computes the log loss summed over all possible classes.
\[\textrm{CCE}_i = - \sum_{\textrm{classes}}y \log(\hat{y})\]
This scoring is computed over all pixels and the average taken. However, this loss function is susceptible to class imbalance. For unbalanced data, training might be dominated by the most prevalent class.

\subsubsection{Dice loss}
The Dice loss function is derived from the Sørensen-Dice Coefficient (see \S\ref{sec:dice}), which is robust to class imbalance as it balances between precision and recall \cite{sudre2017generalised}.

\[(\textrm{Dice loss})_i = 1 - \frac{\sum_{\textrm{classes}} (\textrm{Dice coef.})_\textrm{class}}{\textrm{\# classes}}\]

\subsubsection{Dice + focal loss}
Focal loss \cite{lin2017focal} modifies the pixel-wise cross-entropy loss by down-weighting the loss of easy-to-classify pixels based on a hyperparamter $\gamma$, focusing training on more difficult examples.  Focal loss is defined as:
\[(\textrm{Focal loss})_i = - \sum_{\textrm{classes}}(1 - \hat{y})^{\gamma}y \log(\hat{y}).\]

Dice + focal loss combines the two objectives with a mixing parameter $\alpha$, balancing global (Dice) and local (focal) features of the target mask. Default values of $\gamma=2$ and $\alpha=1$ were used during training.

\section{Experiments}\label{exp-settings}

\practicalfigs

We trained nine image segmentation models for this task, one with each combination of architecture and loss function described above, each on the same randomly selected 49,864 images sampled from our 60,000 image dataset, with a validation set of 5,306 similarly selected images, and a test set of 6,000.  The input and output layers were fixed at a size of 256 $\times$ 256, and input images were normalized using image mean and variance from ImageNet \cite{imagenet}.

A narrow range of learning rates was selected for each experiment by varying the learning rate over a small number of training batches using FastAI's \texttt{lr\_find()} method and selecting a region of greatest descent for the loss.  Once this region was selected, learning rate annealing was used during Adam optimization with otherwise default parameters.  Each model was trained for five epochs, with early stopping at a patience of two; though the loss appeared to plateau in all cases, the early stopping criterion was met in none.

Data augmentation (flip with $p = 0.5$, transpose with $p = 0.5$, rotate with $p = 0.4$) was applied to each batch during training.\footnote{
    Augmentation was performed using Python's \texttt{albumentation} library \cite{albumentations}.
}  Weight decay was set to $1\mathrm{e}{-2}$ and batch normalization was used.

\subsection{Model selection criterion}\label{sec:dice}

We chose the Sørensen-Dice coefficient as a model selection criterion.  The Dice coefficient, equivalent to F1 score, is computed pixel-wise between the predicted and target mask where \[\textrm{Dice coef.} = \frac{2TP}{2TP + FP + FN}.\] For an aggregate measure of model performance, the macro-average Dice coefficient was considered.

\subsection{Unknown target test}

After selecting a model based on this criterion, we evaluated its performance on the 1500 images from the set of images of the `unknown' target, as described in \S\ref{sec:ood}.

\subsection{Practical test}

To evaluate how well a model trained on these data might perform in a practical application, we obtained three photographs of unmanned spacecraft in low Earth orbit taken from the vantage point of another on-orbit spacecraft and annotated them by hand, evaluating the best-performing segmentation model on each.

\section{Results}

\bigresults

\perclassresults


\begin{figure*}[h]
  \centering
  \subfloat[True mask, known target]{\includegraphics[width=0.225\textwidth]{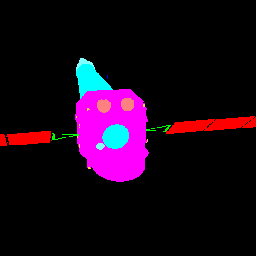}\label{fig:seen-true-2}}
  \hfill
  \subfloat[Predicted mask, known target]{\includegraphics[width=0.225\textwidth]{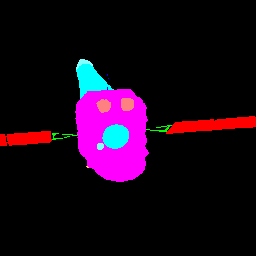}\label{fig:seen-pred-2}}
  \hfill
  \subfloat[True mask, unknown target]{\includegraphics[width=0.225\textwidth]{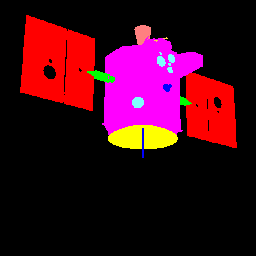}\label{fig:unseen-true-1}}
  \hfill
  \subfloat[Predicted mask, unknown target]{\includegraphics[width=0.225\textwidth]{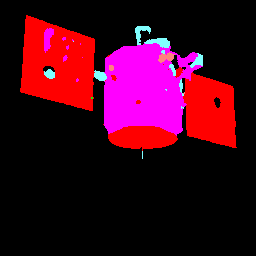}\label{fig:unseen-pred-2}}
  \caption{\textbf{Example out-of-distribution true and predicted masks for a known (Chandra) and unknown (Clementine) target.}}\label{fig:synth-res}
\end{figure*}

Table \ref{tab:big_results} shows Sørensen-Dice coefficients for each trained model on validation, known target test, and unknown target test sets.  Performance on spacecraft present in the training and validation sets is consistently high, however, all of the trained models struggle on the unknown target spacecraft.  The U-Net model with categorical cross-entropy loss showed highest performance on the known target test, with a Sørenson-Dice coefficient of 0.8723; for an unknown target, the DeepLab model with CCE loss was highest at 0.2519.  The gap in performance between known and unknown targets was relatively consistent across methods.  

Figure \ref{fig:synth-res} shows example ground truth and model-predicted masks for the selected U-Net model for a known and unknown target spacecraft.

Table 3 shows per-class Sørensen-Dice coefficients from the selected model on the in-distribution and out-of-distribution test data. 

\subsection{Practical test results}

Figure \ref{fig:pract-test} shows the results of the practical test of the highest performing segmentation model on three photographs, taken from the NASA Image and Video Library, of on-orbit unmanned spacecraft.  The results are more or less consistent with synthetic images of unknown target spacecraft; notably, for the image of the Space Flyer Unit, some sensor or image artefacts, imperceptible to the human eye, resulted in false positive classification of the background and a ~0.05 degradation in Sørenson-Dice coefficient; this could be mitigated in practise with classical object detection methods.

\section{Conclusions}

Benchmark results on these synthetic data show promise for semantic image segmentation in this domain, and show deep architectures for semantic segmentation can learn to recognize many different spacecraft components and categorize them appropriately by type.

Even with a high degree of class imbalance, Dice loss and Dice + focal loss did not always lead to an improvement in model performance, perhaps owing to CCE loss having a smoother gradient than that of Dice loss, resulting in a less noisy descent path during optimization. \cite{anantrasirichai2019defectnet}.

Our unknown target test shows that the selected model struggles to generalize beyond the four main spacecraft in the dataset for nearly every type of spacecraft component; the selected model is able to identify those larger components (main module, solar panels) of the Clementine spacecraft which have a variety of configurations in the training data, but did overfit to smaller components (sensors, thrusters, etc.), possibly by recognizing a limited number of examples of each.  Notably, the selected model mis-classified the parabolic reflector for the Clementine spacecraft almost entirely (Dice $<0.001$); our synthetic data contained only one example of this type of component, on the NEAR spacecraft, which the selected model was able to classify quite well on test data (Dice $\approx .95$).  

Semantic image segmentation for autonomous rendezvous may be achievable with these methods \emph{if the target spacecraft is known}, pending further tests with physical simulation and on-orbit testing.  Image segmentation for arbitrary targets may still be achievable with more representative training data or different methods.  Shallower architectures may merit exploration, since this domain likely has a much smaller collection of salient features, and a smaller semantic space than more general segmentation tasks.  Pre-training on large datasets of unrelated images such as ImageNet may have hindered, rather than helped, the benchmark models to learn the appropriate features.

For the practical test, the observed evaluation metric seemed consistent with that of the unknown target test, suggesting a need to experiment with synthetic images of a target for which real sensor data are available.  Additionally, our data synthesis pipeline did not simulate camera or sensor effects; this is one way in which the data synthesis pipeline presented here could be improved.  

If these limitations can be addressed, a data synthesis pipeline such as this would allow for scaling the number of training data to the complexity of a machine learning task, enabling development of sophisticated perceptual systems \emph{in silico} for applications where acquiring real sensor data at scale is prohibitively expensive.

\acknowledgments
The authors would like to thank Jesse Trutna of Stanford's Space Rendezvous Lab, for inspiring us to think about applications of machine learning to research in autonomous spacecraft operations; Kevin Okseniuk, a program manager at SpaceWorks Enterprises, for generously assisting the authors by helping define the semantic space for our class labels and teaching us how to recognize the components they correspond to; and Laura Rakiewicz, an architectural designer at Alliiance, who assisted in hand-annotation of the practical test photographs.

\bibliographystyle{IEEEtran}
\bibliography{synthetic-semantic}




\thebiography
\begin{biographywithpic}
{William S.~Armstrong}{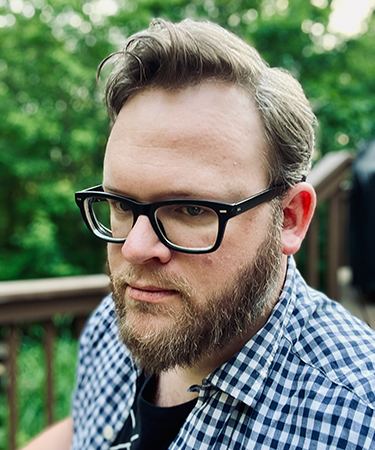}
recieved a B.A.~in Mathematics from University of Minnesota in 2005.  After a decade-long career in mathematical risk modeling for the insurance and financial services sector, he worked as a consultant in data science and machine learning, and now heads the machine learning research and development group at a Fortune 500 retailer.  He is a part-time student with the Stanford Center for Professional Development, and has contributed to research in natural language processing.  He is deeply passionate about astronomy and space exploration.
\end{biographywithpic} 

\begin{biographywithpic}
{Spencer Drakontaidis}{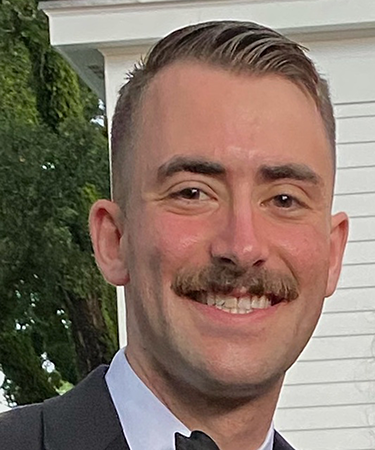}
is a Software Engineer working on tools to help other engineers secure their code more easily and efficiently. He received his B.S.~in Computer Science from the United States Military Academy and a M.S.~in Computer Science from Stanford University. He is passionate about computer security and novel techniques to improve the way society interacts with computing.
\end{biographywithpic}

\begin{biographywithpic}
{Nicholas Lui}{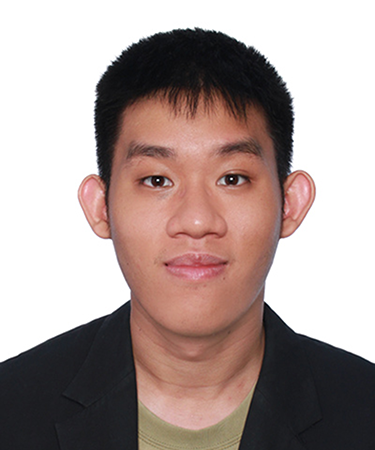}
is a M.S.~in Statistics student at Stanford funded by the Knight-Hennessy Scholars program. He is passionate about the application of tech to social good and has worked on housing equity, criminal justice reform, and climate change. Nicholas received his B.A. degree in Economics from the University of Cambridge.
\end{biographywithpic}

\end{document}